\documentclass[11pt,a4paper]{article}
\usepackage[hyperref]{acl2020}
\usepackage{times}
\usepackage{latexsym}

\usepackage{microtype}

\usepackage{amsfonts}
\usepackage{helvet}
\usepackage{courier}
\usepackage{graphicx}
\usepackage{tikz}
\usepackage{pgfplots}
\usepackage{float}
\usepackage{amsmath}
\usepackage{subcaption}
\usepackage{tikz-dependency}
\usepackage{xcolor}
\usepackage{todonotes}
\definecolor{bblue}{HTML}{4F81BD}
\definecolor{rred}{HTML}{C0504D}
\definecolor{ggreen}{HTML}{9BBB59}
\definecolor{ppurple}{HTML}{9F4C7C}
\usepackage{microtype}
\usepackage{multirow}
\usepackage{amssymb}
\usepackage[utf8]{inputenc} 
\usepackage[T1]{fontenc}
\usepackage{colortbl}
\aclfinalcopy 

\newcommand*{\affaddr}[1]{#1} 
\newcommand*{\affmark}[1][*]{\textsuperscript{#1}}
\newcommand*{\email}[1]{\texttt{#1}}

\title{Mutlitask Learning for Cross-Lingual Transfer of Semantic Dependencies}

\author{%
Maryam Aminian\affmark[1], Mohammad Sadegh Rasooli\affmark[2], Mona Diab\affmark[1]\\
\affaddr{\affmark[1]Department of Computer Science, The George Washington University, Washington, D.C.}\\
\affaddr{\affmark[2]Department of Computer and Information Science,  University of Pennsylvania, Philadelphia, PA }\\
\affaddr{\affmark[1]\email{\{aminian,mtdiab\}@gwu.edu}}, \affaddr{\affmark[2]\email{rasooli@seas.upenn.edu}}\\
}

\date{}

\begin{document}

\maketitle
\begin{abstract}
We describe a method for developing broad-coverage semantic dependency parsers for languages for which no semantically annotated resource is available. We leverage a multitask learning framework coupled with an annotation projection method. We transfer supervised semantic dependency parse annotations from a rich-resource language to a low-resource language through parallel data, and train a semantic parser on projected data. We make use of supervised syntactic parsing as an auxiliary task in a multitask learning framework, and show that with different multitask learning settings, we consistently improve over the single-task baseline. In the setting in which English is the source, and Czech is the target language, our best multitask model improves the labeled F1 score over the single-task baseline by $1.8$ in the in-domain SemEval data~\cite{oepen-etal-2015-semeval}, as well as $2.5$ in the out-of-domain test set. Moreover, we observe that syntactic and semantic dependency direction match is an important factor in improving the results.
\end{abstract}

\section{Introduction\label{sec:sdp:intro}}
Broad-coverage semantic dependency parsing (SDP)\footnote{We use \emph{broad-coverage semantic dependency} and \emph{semantic dependency} interchangeably throughout this paper.}, which is first introduced in the SemEval shared task~\cite{oepen-etal-2014-semeval}, aims to provide the semantic analysis of sentences by capturing the semantic relations between all content-bearing words in a given sentence. This is in contrast to models such as semantic role labeling \cite{baker-etal-1998-berkeley} that focus on verbal and nominal predicates. Compared to other semantic representations~\cite{baker-etal-1998-berkeley,surdeanu-etal-2008-conll,hajivc2009conll,banarescu2013abstract,abend-rappoport-2013-universal}, SDP covers a wide range of semantic phenomena such as negation, comparatives, possessives and various types of modifications that have not been analyzed in other semantic models.
Furthermore, extensive structural similarities between syntactic and semantic dependencies make it easier to adopt existing dependency parsers to have efficient inference. Despite all advantages provided by SDP, resources with annotated semantic dependencies are limited to  languages released in the SemEval shared tasks \citep{oepen-etal-2014-semeval, oepen-etal-2015-semeval, che-etal-2016-semeval}, namely English, Czech and Chinese. This motivates us to develop SDP models for languages without semantically annotated data.

\begin{figure*}[t!]
\begin{center}
    
\scalebox{0.53}{
 \begin{dependency}[theme = simple]
		\begin{deptext}
			 I \& congratulate \& the \& rapporteur \& , \& Mr \& Garriga \& , \& on \& his \& integrated \& report \& , \& which \& sets \& out \& the \& political \& priorities \& of \& a \& different \& Europe \& from \& the \& Europe \& we \& know \& . \\
			\\
			\\
			\\
		\& \&	Blahopřeji \& zpravodaji \& , \& panu \& Garrigovi \& \colorbox{blue!30}{,} \& k \& jeho \& ucelené \& zprávě \& \colorbox{blue!30}{,} \& která \& stanovuje \& politické \& priority \& jiné \& Evropy \& \colorbox{blue!30}{než} \& \colorbox{blue!30}{té} \& \colorbox{blue!30}{,} \& \colorbox{blue!30}{kterou} \& známe \& . \\
		\end{deptext}
	
		\path[dashed,color=blue] (\wordref{1}{2}) edge (\wordref{5}{3});
		\path[dashed,color=blue] (\wordref{1}{29}) edge (\wordref{5}{25});
		\path[dashed,color=blue] (\wordref{1}{4}) edge (\wordref{5}{4});
    \path[dashed,color=blue] (\wordref{1}{12}) edge (\wordref{5}{12});
    \path[dashed,color=blue] (\wordref{1}{5}) edge (\wordref{5}{5});
		\path[dashed,color=blue] (\wordref{1}{9}) edge (\wordref{5}{9});
		\path[dashed,color=blue] (\wordref{1}{19}) edge (\wordref{5}{17});
    \path[dashed,color=blue] (\wordref{1}{11}) edge (\wordref{5}{11});
     \path[dashed,color=blue] (\wordref{1}{28}) edge (\wordref{5}{24});
		\path[dashed,color=blue] (\wordref{1}{7}) edge (\wordref{5}{7});
		\path[dashed,color=blue] (\wordref{1}{10}) edge (\wordref{5}{10});
    \path[dashed,color=blue] (\wordref{1}{15}) edge (\wordref{5}{15});
     \path[dashed,color=blue] (\wordref{1}{18}) edge (\wordref{5}{16});
		\path[dashed,color=blue] (\wordref{1}{23}) edge (\wordref{5}{19});
		\path[dashed,color=blue] (\wordref{1}{6}) edge (\wordref{5}{6});
    \path[dashed,color=blue] (\wordref{1}{14}) edge (\wordref{5}{14});
    \path[dashed,color=blue] (\wordref{1}{22}) edge (\wordref{5}{18});
		

		
 		\deproot{2}{TOP}
 		\depedge{2}{1}{ACT-arg}
 		\depedge{2}{4}{ADDR-arg}
 		\depedge{5}{4}{APPS.member}
 		\depedge{7}{6}{NE}
 		\depedge{2}{7}{ADDR-arg}
 		\depedge{5}{7}{APPS.member}
 		\depedge{12}{9}{APP}
	    \depedge{12}{10}{RSTR}
        \depedge{2}{12}{PAT-arg}
        \depedge{15}{14}{ACT-arg}
        \depedge{12}{15}{RSTR}
        \depedge{19}{18}{RSTR}
         \depedge{15}{19}{PAT-arg}
        \depedge{23}{22}{RSTR}
        \depedge{19}{23}{APP}
        \depedge{28}{27}{ARC-arg}
	   
	   \deproot[dashed,edge below]{3}{TOP}
	   \depedge[dashed,edge below]{5}{4}{APPS-member}
	   \depedge[dashed,edge below]{3}{4}{ADDR-arg}
	   \depedge[dashed,edge below]{7}{6}{NE}
	   \depedge[dashed,edge below]{5}{7}{APPS.member}
	   \depedge[dashed,edge below]{3}{7}{ADDR-arg}
	   \depedge[dashed,edge below]{12}{9}{APP}
	   \depedge[dashed,edge below]{12}{10}{RSTR}
	   \depedge[dashed,edge below]{3}{12}{PAT-arg}
	    \depedge[dashed,edge below]{15}{14}{ACT-arg}
	     \depedge[dashed,edge below]{12}{15}{RSTR}
	      \depedge[dashed,edge below]{17}{16}{RSTR}
	    \depedge[dashed,edge below]{15}{17}{PAT-arg}
	      \depedge[dashed,edge below]{19}{18}{RSTR}  
	      \depedge[dashed,edge below]{17}{19}{APP}
	      
	   \wordgroup[color=purple]{1}{2}{2}{pred}
	   \wordgroup[color=purple]{5}{3}{3}{pred}
	   
	   \wordgroup[color=blue]{1}{5}{5}{pred}
	    \wordgroup[color=blue]{5}{5}{5}{pred}
	    
	    \wordgroup[color=black]{1}{7}{7}{pred}
	    \wordgroup[color=black]{5}{7}{7}{pred}

	    \wordgroup[color=orange]{1}{12}{12}{pred}
	    \wordgroup[color=orange]{5}{12}{12}{pred}

	    \wordgroup[color=green]{1}{15}{15}{pred}
	    \wordgroup[color=green]{5}{15}{15}{pred}

	   \wordgroup[color=red]{1}{19}{19}{pred}
	    \wordgroup[color=red]{5}{17}{17}{pred}
	    
 \wordgroup[color=gray]{1}{23}{23}{pred}
	    \wordgroup[color=gray]{5}{19}{19}{pred}
		
	\end{dependency}
}
\end{center}
\caption{An example of annotation projection for an English-Czech sentence pair from the Europarl corpus~\cite{koehn2005europarl}. Supervised dependencies are shown on top, and projected dependencies, obtained using word alignments (dashed lines), are shown at the bottom.\label{fig:annotation_proj_b}}

\end{figure*}


This paper describes a method for developing a semantic dependency parser for a language without annotated semantic data. We assume that a supervised syntactic dependency parser, as well as translated text (parallel data) are available in the target language of interest. This is indeed a realistic scenario: there are about 90 languages with syntactic annotation in the Universal Dependencies corpus~\cite{11234/1-1699} while we only have a few languages with annotated semantic dependencies~\cite{oepen-etal-2014-semeval,oepen-etal-2015-semeval,che-etal-2016-semeval}. We make use of \emph{annotation projection} for transferring supervised semantic annotations from a rich-resource source language to the target language through word alignment links.  Figure~\ref{fig:annotation_proj_b} shows an example of annotation projection for semantic dependencies. 

We further propose a simple but effective multitask learning framework to leverage supervised syntactic parse information in the target language and improve the representation learning capability in the intermediate layers of the semantic parser. For obtaining such a goal, we benefit from the similarity between syntactic and semantic dependency parsers~\cite{dozat2016deep,dozat-manning-2018-simpler} via sharing intermediate layers. To the best of our knowledge, this work is the first study to develop semantic dependency parsers in the absence of annotated data. Our multitask learning approach, despite its simplicity, yields significant improvements in the performance of the vanilla semantic dependency parser built using annotation projection. We conducted experiments to build semantic dependency parser using projected SDP from English to Czech and compared that with the multitasking setup where we use target syntactic dependencies. Our results show improvements in the labeled F1 from $57.5$ to $59.3$ ($74.3$ to $76.9$ for unlabeled F1) on the SemEval 2015 in-domain data, and even larger gain from $59.0$ to $61.5$ ($75.8$ to $78.5$ for unlabeled F1) on the SemEval 2015 out-of-domain set. We additionally explore the efficacy of contextualized word representations, BERT~\cite{devlin-etal-2019-bert} and ELMO~\cite{peters2018deep} as features in our annotation projection model and find a marginal gain by using those contextual features.

The main contributions of this paper are as the following:
\begin{itemize}
\item We propose an annotation projection approach for syntactic dependencies: to the best of our knowledge, ours is the first work that develops semantic dependency parsers without annotated data.
\item We show that multitask learning is an easy but effective approach for improving the accuracy of semantic parsers. Although there is marginal improvement from multitask learning in a supervised setting, the improvement in annotation projection is promising. We also show that there is enough statistical evidence on the role of syntactic guidance in semantic parsing.
\item We run an extensive set of experiments with and without contextualized word representations, and obtain significant improvements compared to a strong annotation projection baseline.
\end{itemize}

\section{Related Work}

Representing the underlying semantic structure of a sentence through bilexical semantic dependencies as a directed graph have been the subject of many studies \cite{baker-etal-1998-berkeley,surdeanu-etal-2008-conll,banarescu2013abstract,abend-rappoport-2013-universal,oepen-etal-2014-semeval}. After the SemEval shared tasks on broad-coverage semantic dependency parsing~\cite{oepen-etal-2014-semeval,oepen-etal-2015-semeval,che-etal-2016-semeval}, there have been many studies on this topic~\cite{du-etal-2015-peking,che-etal-2016-semeval, chen-etal-2018-neural,almeida-martins-2015-lisbon, wang2018neural, dozat-manning-2018-simpler, kurita2019multi,stanovsky-dagan-2018-semantics}. 
There are other semantic formalisms, all of which intend to produce sentence-level semantic analysis by generating a graph covering content words in sentence such as the Universal Conceptual Cognitive Annotation (UCCA) \citep{abend-rappoport-2013-universal}, Abstract Meaning Representation (AMR) \citep{banarescu-etal-2013-abstract}, frame semantics~\cite{baker-etal-1998-berkeley}, and dependency-based semantic roles~\cite{surdeanu-etal-2008-conll}.

 The SemEval dataset consists of three main semantic representations, also known as target representations, that are slightly different in the way they determine and label semantic dependencies. These representations are usually referred to as {DM} (DELPH-IN minimal recursion), {PSD} (Prague semantic dependencies) and {PAS} (Enju predicate–argument structures).
 Since SemEval Czech training and evaluation data uses PSD target representation, we choose the PSD representation for English as well.
 Motivated by the fact that each of these representations cover different aspects of sentence-level semantics, there has been a line of study to use multitask learning in order to improve single-task baselines~\cite{peng-etal-2017-deep,peng-etal-2018-learning,hershcovich-etal-2018-multitask,kurita2019multi}. For example, \newcite{peng-etal-2017-deep} improve a single-task SDP baseline by exploiting data provided for different SDP representations. \newcite{peng-etal-2018-learning} introduce a multitask system that is trained on disjoint semantic datasets. In another work, \newcite{hershcovich-etal-2018-multitask} applies multitask learning over different semantic annotation formalisms including UCCA, AMR, and DM as well as universal dependencies. In a more recent study, \newcite{kurita2019multi} describe a SDP model that combines the arc-factored and graph-based parsing algorithms. 

The role of syntax in a multitasking framework has been studied in other semantic parsing tasks such as semantic role labeling \cite{lluis-etal-2013-joint,swayamdipta-etal-2018-syntactic}. \newcite{lluis-etal-2013-joint} describe a joint arc-factored parser that uses dual decomposition for jointly decoding syntactic and semantic structures. In another work, \newcite{swayamdipta-etal-2018-syntactic} introduce a method for discovering semantic structure by incorporating syntactic information in a multitasking framework. 

Annotation projection~\cite{yarowsky2001inducing} has been extensively used in other tasks such as part-of-speech tagging~\cite{tackstrom2013token}, named entity recognition~\cite{ehrmann2011building}, dependency parsing~\cite{hwa2005bootstrapping,mcdonald-petrov-hall:2011:EMNLP,cohen-das-smith:2011:EMNLP}, semantic role labeling~\cite{pado_projection,akbik_projection}, AMR parsing~\cite{damonte-cohen-2018-cross}, and sentiment analysis~\cite{mihalcea2007learning}. This paper is the first work that aims to build a SDP model using cross-lingual transfer systems without any annotations in the target language of interest. We also consider integrating syntactic inductive bias in a multitasking framework in any cross-lingual SDP models.

\section{Approach}
In this section, we briefly describe the problem, and then describe our proposed method.

\subsection{The Parsing Model}
\label{sec:parsing_model}
For an input sentence $x=x_1, \cdots, x_n$ that has $n$ words, the goal of a semantic parsing model is to learn binary dependency decisions $y_{i,j} \in \{0,1\}$ for every head index $0\leq i\leq n$ and dependent index $1 \leq j \leq n$, in which the zeroth index ($x_0$) indicates the dummy \emph{root} token. For every head-dependent pair $(i,j)$, such that $y_{i,j}=1$, the parser should find a label $l_{i,j}$ from a set of predefined semantic dependency labels $\cal L$. In most cases, the parsing decision is decomposed in two decisions: unlabeled dependency parsing, and labeling each dependency edge. The only constraint here is that the final semantic graph should be acyclic.

We use the standard model of \newcite{dozat-manning-2018-simpler} for which the parsing model is based on a simple head selection algorithm. This model learns dependency edge scores $s^{\tt edge}(i,j)$ for all possible head-dependent pairs $(i, j)$. The final parsing decision is basically a sign function:
\[
y_{i,j} = \{ s^{\tt edge}(i,j) \geq 0 \}
\]
Similarly, the parser learns a labeling function $s_{\tt l}^{\tt label}(i,j)$ for every pair that $y_{i,j}=1$:
\[
l_{i,j}= \arg\max_{{\tt l} \in \cal L}  s_l^{\tt label}(i,j)
\]

The model of \newcite{dozat-manning-2018-simpler} uses a deep model for which the first layer is the embedding layer that consists of word, part-of-speech tag, and character representations. The second layer consists of deep bidirectional LSTMs~\cite{hochreiter1997long} that construct recurrent representations $r_i$ for every word position $i$. The third layer uses four single-layer feed-forward neural networks (FNN) as attention mechanisms one for heads and one for dependents both for dependency labels and dependency edges (four FNNs for four different attentions):

\[
h_i^{\tt type} = {\tt FNN}^{\tt  type}(r_i)
\]
where ${\tt type}$ is in one of the four possibilities: \emph {edge-dep}, \emph {edge-head}, \emph{label-dep}, and \emph{label-head}.

The final layer uses the FNN output in either a bilinear function:
\[
s^{\tt edge}(i,j)= f^{\tt edge}(h_i^{\tt edge-dep}, h_i^{\tt edge-head})
\]
and
\[
s^{\tt label}(i,j)= f^{\tt label}(h_i^{\tt label-dep}, h_i^{\tt label-head})
\]


During training, the Sigmoid cross-entropy function is used for the edges, and the softmax cross-entropy function is used for the labels. The two losses are interpolated to calculate the final loss value with a coefficient $0<\lambda<1$.

\subsection{Annotation Projection}
The main idea behind annotation projection is to project supervised annotations from a rich-resource language to a target language through alignment links. For a source sentence  $x'=x'_1, \cdots, x'_m$ with $m$ words, and a target sentence $x=x_1, \cdots, x_n$ with $n$ words, we achieve one-to-one alignments by running an unsupervised word alignment algorithm (in our case, Giza++~\cite{och2000giza}) on both directions. We use the intersected alignments $a=a_1, \cdots, a_m$ such that $0 \leq a_i\leq n$ where $a_i=0$ indicates a {\tt null} or empty alignment. For every source dependency relation $y'_{i, j} \in \{0,1\}$ where $a_i,a_j \neq 0$, we project the dependency to the target sentence $y_{a_i, a_j}=y'_{i, j}$ (if $i=0$ then $a_i=0$). Therefore, we can project the dependency label as well $l_{a_i, a_j}=l'_{i, j}$. 

Due to inevitable missing alignments, some $y_{i,j}$ might  remain undecided leading to partial assignments to the set $y$. In practice, this happens in many sentences mainly due to reasons including lack of lexical correspondence between the source and target, translation shift, and alignment errors. Thanks to the flexibility of deep learning models, we can cancel back-propagation for those undecided directions by appropriate masking. The rest of the training procedure remains the same: we then train a supervised parsing model on projected dependencies despite the fact that the dependencies are not gold standard. Unfortunately this is not ideal since the projected dependencies are not always correct, but we can still tailor them for our case in which no gold standard annotated data is available.

\subsection{Multitask Learning with Syntactic Dependencies}
The main goal of multitask learning is benefiting from structural or statistical similarities found in one or more auxiliary tasks to improve the model learned for the target task~\cite{caruana1997multitask}. Often times, auxiliary tasks are chosen amongst closely related tasks. We consider syntactic dependency parsing as an auxiliary task for semantic dependency parsing. These two tasks have many similarities both in terms of unlabeled edges, and some correlations between labeled edges. 

We parse the target side of parallel data with our supervised syntactic parser.\footnote{Based on preliminary experiments, using projected syntactic dependencies do not improve the accuracy of the semantic parsing model.} Our training data for the target language has partial semantic annotations plus fully parsed supervised syntactic trees. There are of course many differences, and sometimes completely different dependency directions (such as the punctuation directions in Figure~\ref{fig:annotation_proj_b}).  However, syntactic dependencies can be mostly helpful for sentences for which the number of projected dependencies are very small. In general, we expect multitask learning to help better represent intermediate layers of the semantic parser. 

The architecture of syntactic dependency parsers is very similar to that of semantic parsers. The main difference lies in the fact that each word can only have one head: if $y_{i,j}=1$ for dependent index $j$, then $y_{k,j}$ for all $k\neq i$ should be zero. The syntactic dependency parser of \newcite{dozat2016deep} is very similar to its semantic dependency parser counterpart. In syntactic parsing, instead of using a sign function, we take the highest scoring edge as a head selection algorithm. Other components are similar to those described in \S\ref{sec:parsing_model} for the semantic parser. 
The overall loss value for the multitask model is computed by interpolating semantic and syntactic losses using an interpolation coefficient $\omega$.

\begin{figure}[t]
    \centering
        \includegraphics[width=0.5\textwidth,height=6cm]{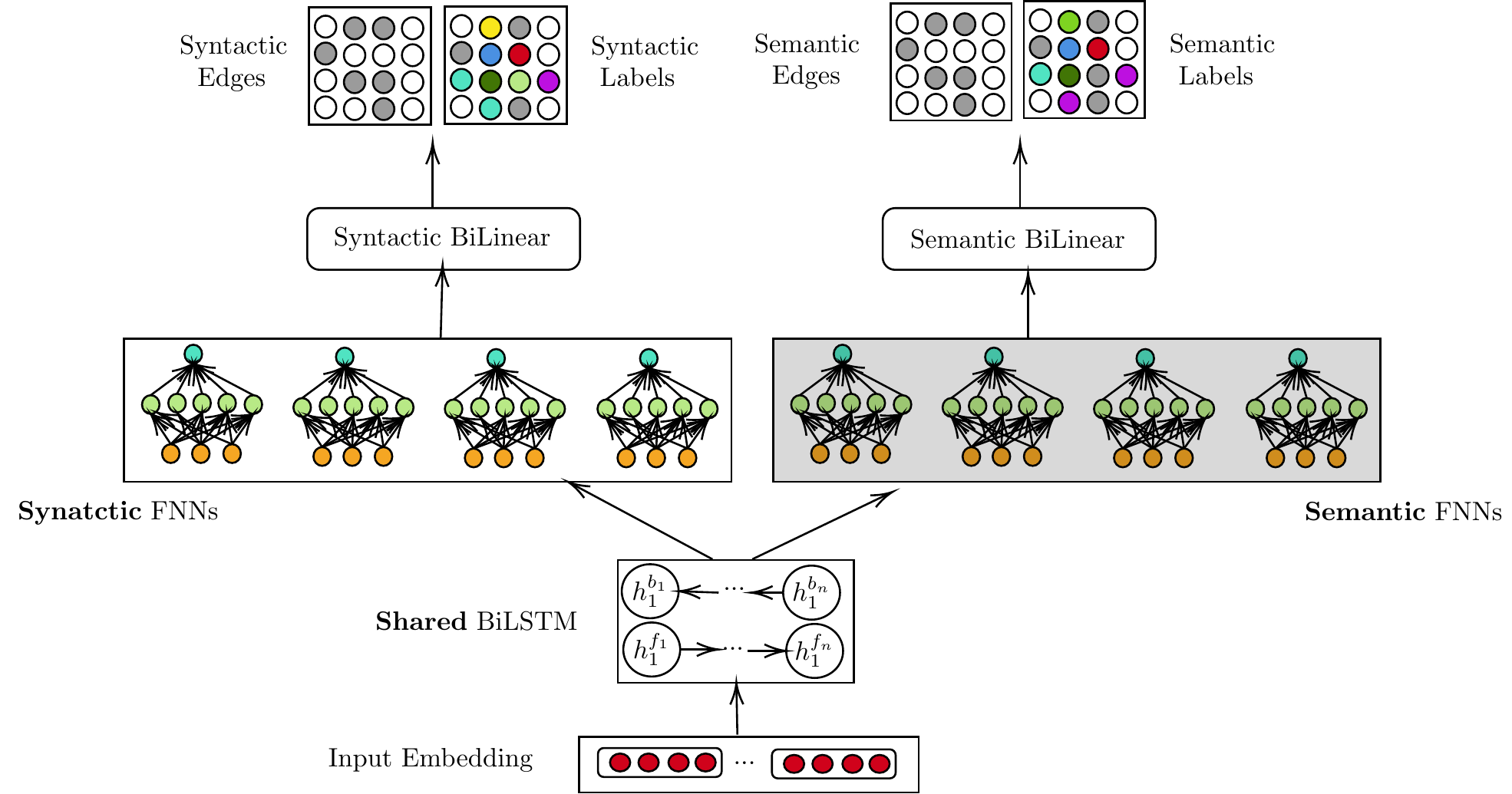}
    \caption{A multitask architecture for semantic and syntactic dependency parsing where the embedding and recurrent layers are shared across the two tasks. This architecture gives us the best results in most cases.}
    \label{fig_sdp_mtl_sharedFNN}
\end{figure}


Here we can share any part of the deep model except the final task-specific layers. We make use of \emph{projected semantic dependencies} as well as \emph{supervised syntactic dependency parses} during training the annotation projection model. In order to find out the exact parameter sharing structure that leads to the best SDP model for the target language, we try the following options: 1) Only sharing the first two layers (embedding, and recurrent layer), 2) Sharing the first two layers, but with an additional task-specific recurrent layer, 3) Sharing the all three layers, but with an additional task-specific recurrent layer, and 4) Sharing all intermediate layers. Figure~\ref{fig_sdp_mtl_sharedFNN} shows the first case for which only the first two layers are shared between the two tasks. 

\section{Experiments and Results\label{sec:sdp:experiments}}
In this section, we briefly describe the setting, and show the results.

\subsection{Data and Setting}
We consider English as the source language and Czech as the target language. We use the SemEval 2015~\cite{oepen-etal-2015-semeval} test data sets to evaluate our models.\footnote{Due to tokenization inconsistencies, we were not able to run our experiments on Chinese data.} Since the PSD annotation is available in both English and Czech, we use that throughout all of our experiments. We use Giza++~\cite{och2000giza} with its default configuration to obtain word alignments on the Europarl parallel corpus \citep{koehn2005europarl} by running it in both source-to-target and target-to-source directions, and extracting intersected word alignments. 

We perform evaluations against both in-domain and out-of-domain test sets provided by SemEval 2015 shared task \cite{oepen-etal-2015-semeval}. All evaluations are performed using the official scoring script provided by the shared task. We compare the multitask results with the Single baseline model. Our preliminary experiments show that using the full set of projections (612k sentences) to train the model yields equivalent performance to the model trained on a sample with a reasonable size (80K in this case). We train all of our models, including the single-task baseline, on a random sample of 80K sentences from the projected corpus. We sample our data based on the density of alignments, which is the number of aligned words divided by number of words in a sentence. We use a median $0.8$ for a sample containing 80,000 sentence where half of sentences have a density less than 0.8 and the other half comprise sentences whose projection density is more than $0.8$. In order to simulate a fully unsupervised approach, we use $5\%$ of the projected data as the held-out data to decide when to stop training.


\begin{table*}[t!]
    \centering
    \small
    \begin{tabular}{l|| c c   | c | c | c c | c c | c c || c c }
    \hline \hline
      \multirow{2}{*}{Model} & \multicolumn{2}{c}{Shared} &Task & In-domain &  \multicolumn{2}{c}{Transfer}& \multicolumn{2}{c}{+ ELMO}  & \multicolumn{2}{c}{+ mBERT}  & \multicolumn{2}{c}{Supervised} \\ 
 &   RNN & FNN &  RNN & data & LF & UF & LF & UF & LF & UF & LF & UF \\ \hline\hline
 \multirow{2}{*}{Single} & & & & \checkmark & 57.5 & 74.3 & 57.4 & 74.5 & 56.3 & 75.0 & 85.4 & 91.1 \\
 & & & &  \cellcolor{green!7}   & \cellcolor{green!7} 58.8 & \cellcolor{green!7} 75.8 & \cellcolor{green!7} 59.0 & \cellcolor{green!7} 75.8 &  \cellcolor{green!7} 57.9 & \cellcolor{green!7} 76.3  & \cellcolor{green!7} 70.4 & \cellcolor{green!7} 86.8 \\\hline 
 \multirow{8}{*}{Mutitask}  &  \multirow{2}{*}{\checkmark}  &  \multirow{2}{*}{\checkmark} & &  \checkmark & {\bf 59.3} & 76.4 & {\bf 59.3} & 76.9 & 58.3 & 77.0 & 85.1 & 90.1 \\
   & & & & \cellcolor{green!7} & \cellcolor{green!7} 61.2 & \cellcolor{green!7} 78.2 & \cellcolor{green!7} {\bf 61.5} & \cellcolor{green!7} 78.2  &  \cellcolor{green!7} 60.6 & \cellcolor{green!7} {\bf 78.6}    & \cellcolor{green!7} 70.8 &\cellcolor{green!7} 87.0 \\ \cline{2-13}
 & \multirow{2}{*}{\checkmark}  &  \multirow{2}{*}{\checkmark} & \multirow{2}{*}{\checkmark}  &\checkmark  & 58.3 & 75.7 & {\bf 59.3} & 76.7 & 58.4 & {\bf 77.1}\textbf{} & 85.1 & 91.0 \\ 
 & & & &\cellcolor{green!7} & \cellcolor{green!7} 60.5 & \cellcolor{green!7} 77.9 & \cellcolor{green!7} 61.4 & \cellcolor{green!7} 78.5  & \cellcolor{green!7} 60.4 & \cellcolor{green!7} {\bf 78.6}  & \cellcolor{green!7} 70.8 & \cellcolor{green!7} 87.3 \\  \cline{2-13}
 & \multirow{2}{*}{\checkmark}  & & & \checkmark & 57.7 & 75.2 &  59.2 & 75.7  & 55.7 & 75.0 & 84.3 & 90.2 \\ 
  & & & &  \cellcolor{green!7}  & \cellcolor{green!7} 59.9 & \cellcolor{green!7} 77.1 & \cellcolor{green!7} 61.4 & \cellcolor{green!7} 77.5  &  \cellcolor{green!7} 58.2 & \cellcolor{green!7} 76.6   & \cellcolor{green!7} 69.7 & \cellcolor{green!7} 86.4 \\  \cline{2-13}
  & \multirow{2}{*}{\checkmark}  & & \multirow{2}{*}{\checkmark}  &  \checkmark & 58.7 & 75.3 & 58.6 & 76.0 & 57.6 & 76.4 & 85.4 & 91.0 \\ 
  & & & & \cellcolor{green!7}  & \cellcolor{green!7} 60.8 & \cellcolor{green!7} 77.3 & \cellcolor{green!7} 60.6 & \cellcolor{green!7} 77.5  &  \cellcolor{green!7} 59.9 & \cellcolor{green!7} 78.1   &  \cellcolor{green!7} 71.1 & \cellcolor{green!7} 87.3 \\\hline\hline
    \end{tabular}

    \caption{Results on the Czech SemEval test data~\cite{oepen-etal-2015-semeval} with different settings. LF and UF denote Labeled and Unlabeled F1 respectively. The first result column (Transfer) does not use contextualized word embeddings. \emph{Task RNN} refers to an extra task-specific embedding in the multitask setting. The shaded rows (even rows) are for out-of-domain data while the other rows are for in-domain data.}\label{tab:id_results}
    \end{table*}

\paragraph{Parsing Parameters}
We use the structural skip-gram model of \newcite{ling-EtAl:2015:NAACL-HLT} for English word embeddings, and run word2Vec~\cite{word2vec} on Wikipedia text to acquire the word vectors for Czech. We use UDpipe \cite{udpipe:2017} pretrained models to produce automatic part-of-speech tags. We train the biaffine dependency parser of \newcite{dozat2016deep} on the Universal Dependencies corpus~\cite{11234/1-1699} to generate supervised syntactic parses in our multitask learning experiments. We mainly use the hyper-parameters used in \newcite{dozat-manning-2018-simpler} except that we use a character BiLSTM without any linear transformation layers\footnote{During development experiments, we found out that linear transformation of characters does not play a significant role in the performance, thereby we excluded this part for simplicity.}. All of our implementations are done using the Dynet library~\cite{neubig2017dynet}.

We use word and part-of-speech vectors of size 100, with 3-layer LSTMs of size 600, and feed-forward layers of size 600. We use a dropout of probability of $0.2$ for words and part-of-speech tags, and $0.25$ for the recurrent and unlabeled feed-forward layers, and $0.33$ for the labeled feed-forward layers. The interpolation constants $\lambda$ and $\omega$ are set to $0.025$ and $0.975$ respectively to prioritize the semantic task as our main task in multitasking framework. We use the Adam optimizer~\cite{kingma2014adam} with a learning rate of $0.001$ on  minibatches of approximately thousand tokens. We also concatenate the contextual vectors to the input layer as additional features to the parser.\footnote{We have tried the contextualized vectors in the recurrent layer by concatenating them to the recurrent output, but we have not seen any gain from using them at that position.} 
We use the pretrained ELMO embeddings~\cite{peters2018deep} of size $1024$ from \cite{che-EtAl:2018:K18-2,fares-EtAl:2017:NoDaLiDa}. We use the pretrained multilingual BERT models~\cite{devlin-etal-2019-bert} of size $768$ from~\newcite{xiao2018bertservice} with 12 layers and 12 heads. Due to computational limitations, we only use the pretrained BERT models in the input layer without finetuning them.

\subsection{Results\label{sec:sdp:results}} 

Table~\ref{tab:id_results} shows the results on in-domain and out-of-domain data with and without contextual word embeddings. The Single row in this table shows the baseline where we use Czech projection data to train the model. The Multitask rows show the results when we aggregate Syntactic parses in our model too. Comparing the labeled F1 scores for different multitask models, we observe that all multitask models outperform the single-task baseline, regardless of the architecture used to train the model. The only exception is in using the BERT features, which might be due to lack of finetuning the BERT vectors. We observe that multitask models yield a larger increase on the out-of-domain test set compared to the in-domain test set. This illustrates the particular power of multitask learning to improve the target semantic parsing model in truly low-resource settings where in-domain training data might not be available. As we see in the results, the multitask model with a shared recurrent layer slightly outperforms other models. We also observe that using a task-specific recurrent layer leads to less improvement compared to other multitask equivalents probably due over-parameterization of the model. We also see marginal gains from using the ELMO embeddings, and some gain in unlabeled score in using BERT without seeing improvement in labeled accuracy. For example, the best model with ELMO has an equal labeled F1 score as the best model without ELMO ($59.3$), however, the unlabeled results improve ($78.6$ vs. $78.2$ in the best BERT vs best base model) in both BERT and ELMO model.

Comparing to a supervised model, we observe that our models are able to get closer to the supervised accuracy in out-of-domain data. There is still a gap in our best performing model ($59.3$) compared to a supervised parser ($85.4$), however, this gap is smaller in out-of-domain data ($61.5$ vs. $71.1$). Our experiments show that multitasking does not lead to noticeable improvements in supervised setups. Based on our observations, a particular characteristic of the PSD target representation might be a reason: some of the semantic edges in the PSD representation are in the opposite direction when compared to the syntactic dependency existing between those words. For instance, in the PSD labeling scheme, the word \emph{position} in the noun clause ``\emph{the position}'' is labeled as the child of the determiner \emph{the} with the SDP label \emph{det--ARG1} while the syntactic edge between these two words is in the exact opposite direction. 
These inconsistencies might lead to a confusion for the model.
Unlike supervised parsing, in annotation projection, we suffer less from this inconsistency: the amount of noise in semantic projections is already high, so that any additional information can be beneficial. 
We also observe that having separate feed-forward layers per task can modestly help tackling this inconsistency between the two tasks.

\begin{figure}[t!]
\centering

\scalebox{0.7}{
\begin{tikzpicture}
\pgfplotsset{small,samples=20,every tick label/.append style={font=\footnotesize}}

    \begin{axis}[
        width  =0.65*\textwidth,
        height = 7cm,
        major x tick style = transparent,
        ybar=2*\pgflinewidth,
        bar width=9pt,
        ylabel = {Labeled precision},
        xlabel = {Dependency length},
        y label style={at={(0.05,0.5)}},
        symbolic x coords={1,2,3,4,5-9,$\ge$10},
        xtick = data,
        scaled y ticks = false,
        ymin=45,
        legend cell align=left,
        legend style={
                at={(1.1,1.1)},
                column sep=1ex,
                font=\footnotesize
        },
    	point meta=explicit symbolic,
    	every node near coord/.append style={font=\scriptsize},
    	nodes near coords align={vertical},
    ]
        \addplot[style={black,fill=white,mark=none,pattern=dots}]
            coordinates {(1,92.15) (2,84.23) (3,81.98) (4,79.48) (5-9,77.85) ($\ge$10,78.24) };
         \addplot[style={black,fill=yellow,mark=none}]   
            coordinates {(1,67.83) (2,57.49) (3,59.83) (4,62.9) (5-9,65.58) ($\ge$10,73.96) };
         \addplot[
         nodes near coords,
         style={black,fill=blue,mark=none}
         ]
            coordinates {(1,67.57)[-0.4\%] (2,58.58)[1.8\%] (3,59.76)[-0.1\%] (4,61.19)[-2.7\%] (5-9,66.75)[1.8\%] ($\ge$10,75.94)[2.7\%]};

        \legend{Supervised,Single, MTL(RNN)}
    \end{axis}
\end{tikzpicture}
}
\caption{Labeled precision of the best-performing multitask model compared to the single-task and supervised models for different dependency lengths.}\label{sdp_dd_edge_len}
\end{figure}
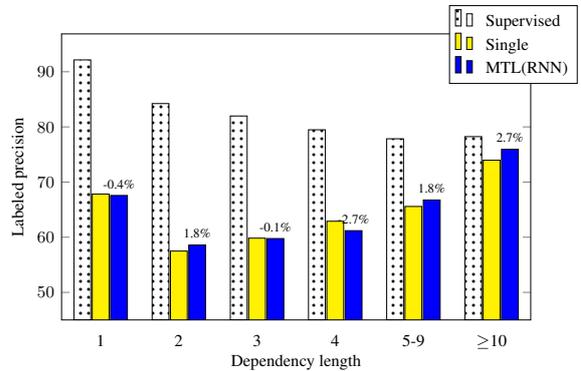

\begin{figure*}[t!]
\begin{center}
\scalebox{1.0}{
\begin{subfigure}{0.5\textwidth}
        \centering
          \input{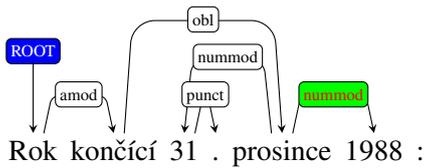}
    \caption{Supervised syntactic tree.}
\end{subfigure}
 \begin{subfigure}{0.5\textwidth}
    \input{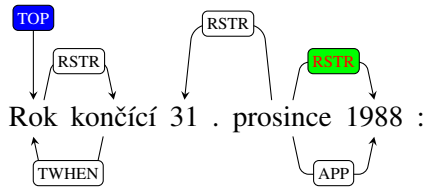}
    \caption{Correct multitask semantic tree (above), and wrong single-task predictions (bottom).}
       \end{subfigure}
}
\end{center}
\caption{An example of a correct semantic tree predicted by a multitask model: the syntactic heads and labels for the first and last words helped the multitask model predict the correct semantic dependency, while the single-task model fails to do so.}\label{fig_syn_match}

\end{figure*}

\begin{figure}[t!]
\centering
\scalebox{0.9}{

\pgfplotsset{compat=1.9}

\begin{tikzpicture}


\begin{axis}[
    ybar stacked,
	bar width=3.5pt,
    legend cell align=left,
        legend style={
                at={(0.95,0.95)},
                column sep=1ex,
                font=\footnotesize},
    ylabel={\% of contribution in improvement},
    ylabel style={font=\footnotesize},
    xlabel={Syntactic dependency relation},
    symbolic x coords={nmod,nsubj,obl,conj,amod,obj,advmod,flat,root,advcl,advmod:emph,acl,ccomp,xcomp,nummod,nummod:gov,obl:arg,det,cc,iobj,
    },
    xtick=data,
    x tick label style={rotate=90,anchor=east, font=\scriptsize},
    totals/.style={nodes near coords align={anchor=south}},
    ]
    
\addplot[blue, fill=blue] plot coordinates {
(cc,0.875273522976)
(ccomp,1.64113785558)
(conj,8.53391684902)
(acl,1.69584245077)
(advmod:emph,1.75054704595)
(advcl,2.84463894967)
(nummod:gov,1.31291028446)
(flat,3.22757111597)
(obl:arg,1.14879649891)
(nsubj,12.8008752735)
(nummod,1.31291028446)
(advmod,7.05689277899)
(xcomp,1.53172866521)
(obl,11.4879649891)
(obj,7.05689277899)
(det,1.03938730853)
(nmod,16.0284463895)
(amod,7.87746170678)
(iobj,0.929978118162)
(root,3.06345733042)
};

\end{axis}
\end{tikzpicture}
}
\caption{Contribution of each syntactic dependency label on improving the semantic dependencies over single-task learning in our best performing multitask model. \label{fig:sdp:analysis:improvements}}

\end{figure}
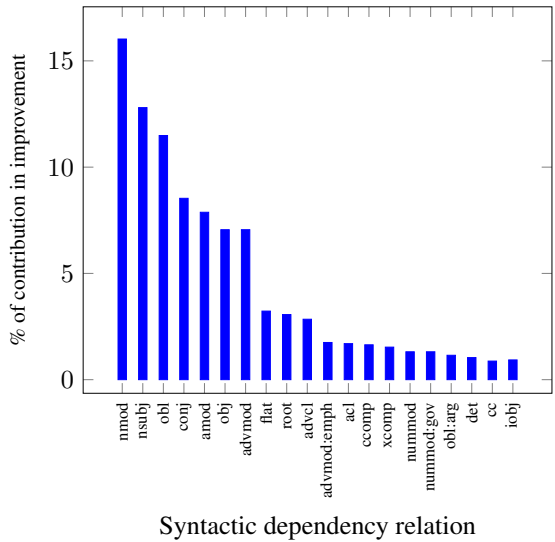

\section{Analysis\label{sec:sdp:analysis}}
In this section, we provide further analysis on the results in order to see how multitask learning improves over the single-task baseline.

\subsection{Effect of Dependency Lengths}
We analyze the performance of our best performing multitask model on different semantic dependencies. Figure \ref{sdp_dd_edge_len} illustrates labeled precision of the model with a shared embedding and recurrent layer compared to the single-task and supervised models for different semantic dependency lengths. We define dependency length as the number of tokens located between a semantic head and its dependent. Numbers shown above each plot denotes the improvement resulting from the multitask model compared to the single-task model. Interestingly, we observe that the multitask model yields relatively more improvements in the longer semantic dependencies compared to the shorter ones, such that the multitask precision for semantic dependencies with length $\ge$10 is noticeably close to the supervised results.  

\subsection{Effect of Syntactic and Semantic Dependency Consistency}
We have done an analysis on the development set for investigating the improvements according to the consistency between syntactic and semantic dependencies. We define a \emph{syntactic head mismatch} as a case for which the gold semantic dependency is in the opposite direction of a syntactic dependency. A syntactic head match is defined as a case in which the semantic and syntactic head of a word is the same. 

We find that there are more syntactic head matches when there is improvement from multitask learning compared to cases for which the single-task model outperforms the multitask model ($60\%$ vs. $49\%$ in unlabeled accuracy, and $54\%$ vs $44\%$ in labeled accuracy). On the other hand, we see less syntactic head mismatch in cases of improvement from multitask model compared to cases of improvement from the single-task model ($6\%$ vs. $8\%$ in unlabeled accuracy, and $7\%$ vs. $10\%$ in labeled accuracy). This is indeed a clear sign that a head match between syntactic and semantic dependencies is a very strong signal for guiding the semantic parser to make better predictions. Figure~\ref{fig_syn_match} shows a simple example extracted from the development data. In this particular example, the two syntactic heads (\emph{ROOT} for the first word, and \emph{nummod} for the sixth word) help the semantic dependency parser discover the correct semantic dependency.



Figure~\ref{fig:sdp:analysis:improvements} shows the percentage of improved semantic dependencies obtained from the best performing multitask setup compared to the single-task model. As shown in this Figure, some syntactic relations yield considerable improvements in semantic dependencies including \emph{nmod}, \emph{nsubj}, \emph{obl}, and \emph{conj}. The sources of this improvement is two-folds: first, there is a direct correspondence between the syntactic and semantic roles in some dependencies such as \emph{conj} and \emph{iobj}, thereby, injecting information about these relations enhances predictions made by the SDP model. Second, the high frequency of relations such as \emph{nmod}, \emph{nsubj}, \emph{obj} and \emph{amod}  might help improve the representation of the learned deep model. 

%


\section{Conclusion}
In this paper, we have described a cross-lingual semantic dependency parsing model based on annotation projection that do not use any annotated semantic data in the target language. We enhance the target semantic model by incorporating syntax in a multitask learning framework. We have shown that our multitasking model outperforms the baseline single-task model on both in-domain and out-of-domain test sets for the Czech language. We further explore density-driven training of the target semantic model as a form of semi-supervised training. Our detailed analysis show that the proposed multitasking approach yields  significant improvements for long distance semantic dependencies.  Future work should investigate the possibility of extending this work on other languages.


\bibliographystyle{acl_natbib}
\bibliography{aaai}

\end{document}